\documentclass[journal]{IEEEtran}
\usepackage{graphicx}
\usepackage[english]{babel}
\usepackage[sort,compress]{cite}
\usepackage{amsmath}
\usepackage{bm}
\usepackage{amsfonts}
\usepackage{subfig}
\usepackage{tikz}
\usepackage{verbatim}

\usepackage{array}
\newcolumntype{L}[1]{>{\raggedright\let\newline\\\arraybackslash\hspace{0pt}}m{#1}}
\newcolumntype{C}[1]{>{\centering\let\newline\\\arraybackslash\hspace{0pt}}m{#1}}
\newcolumntype{R}[1]{>{\raggedleft\let\newline\\\arraybackslash\hspace{0pt}}m{#1}}

\begin{document}

\markboth{Author's Name}{Paper Title}

\title{Did Evolution get it right?\\
An evaluation of Near-Infrared imaging in semantic scene segmentation}

\author{J. Rafid Siddiqui}

\maketitle

\thispagestyle{empty}
\pagestyle{empty}

\begin{abstract}
Animals have evolved to restrict their sensing capabilities to certain region of electromagnetic spectrum. This is surprisingly a very narrow band on a vast scale which makes one think if there is a systematic bias underlying such selective filtration. The situation becomes even more intriguing when we find a sharp cutoff point at Near-infrared point whereby almost all animal vision systems seem to have a lower bound. This brings us to an interesting question: did evolution "intentionally" performed such a restriction in order to evolve higher visual cognition? In this work this question is addressed by experimenting with Near-infrared images for their potential applicability in higher visual processing such as semantic segmentation. A modified version of Fully Convolutional Networks are trained on NIR images and RGB images respectively and compared for their respective effectiveness in the wake of semantic segmentation. The results from the experiments show that visible part of the spectrum alone is sufficient for the robust semantic segmentation of the indoor as well as outdoor scenes.

\end{abstract}

\section{INTRODUCTION}

\vspace{3mm}

\begin{figure}[ht]
      \centering
      \includegraphics[width=\columnwidth]{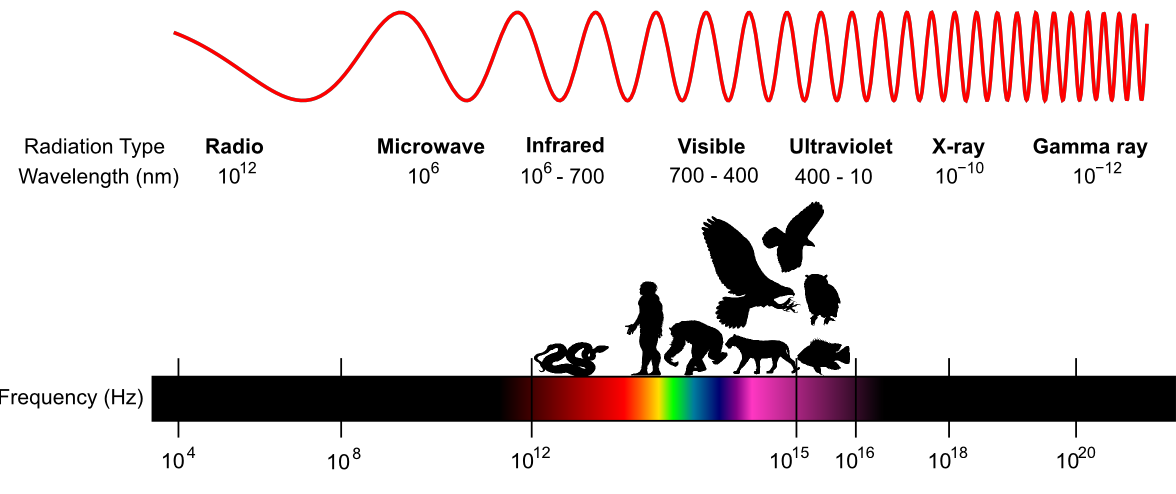}
      \caption{Animals on electromagnetic spectrum based on their vision system sensing capability.}
      \label{fig_animals_spectrum}
\end{figure}

When we look at human vision system in relation to electromagnetic spectrum one thing becomes obvious that humans use merely a fraction of it. It is a common knowledge that humans use only visible part of the spectrum however it seems bizarre when we start to probe into the matter. Why would evolution impose such a restriction on humans that otherwise have been evolved with exceptional brain power in comparison to other fellow animals? One common answer to this question would be that since the process of evolution is a random phenomenon and the fact that it doesn't seek any optimum point therefore, it could just be attributed to mere chance. If that is the case, and considering that evolution is exploratory in nature, one must expect a spread of animals when placed over electromagnetic spectrum. However, reality is rather different - somehow animal visions have evolved to restrict themselves towards mid to higher frequency range and strictly contained in a certain region of the spectrum (Figure \ref{fig_animals_spectrum}). More specifically, if we look closely and search for animals which possess Near-infrared vision, it would soon become clear how hard it is to find such animals. The mammals which have more evolved vision system clearly show a bias towards visible part of the spectrum and have either bi-chromatic or tri-chromatic vision \cite{mammalVision}. One must argue here that there is no correlation between having bigger brains and the kind of sensory capabilities. Then it means we must be able to find an example in birds or reptiles. Most birds as we know today use four color channels for vision processing - the extra channel extends towards higher frequency and shorter wavelength part of the spectrum (i.e. ultraviolet)  \cite{birdVision1}\cite{birdVision2}. Most reptiles stay in visible spectrum except a few. The only one that seems to possess the infrared sensing ability is viper snake \cite{snakeVision1}\cite{snakeVision2}. It is however to be noted that even though these snakes posses the capability to utilize the infrared part of the spectrum using a pit at their forehead, however, this functionality is not part of vision system rather an extension of rudimentary somatosensory sensing system. This brings us to a very interesting point and one must ask here: Is there something at play here? Could it be the case that evolution has intentionally constrained itself towards certain direction in the electromagnetic spectrum thus favoring certain features to evolve such as maybe high level visual processing? 
\par

Since we are interested in finding a relation between the high level visual processing and usage of electromagnetic spectrum it might be useful to mention what it is  meant by it. The high level vision system consists of anatomy and processes which allow an animal to perceive meaningful abstractions from the raw sensory data. Some examples of such functions would be semantic segmentation and scene understanding. Since this functionality is rather rare and is often found in animals which are perhaps more evolved such as primates. This brings us to an interesting question: Does limiting sensory data by restricting access to a certain part of spectrum helps in high level visual cognition such as semantic segmentation? As pointed in \cite{rsi_icra16} that it is perhaps the perceptual filtration at work which makes human better at high level cognitive tasks, it is interesting to analyze this phenomenon further and see whether prioritizing the sensory input indirectly helps solve high level visual cognition. In this work we aim to address this question by performing a series of experiments on Near Infra-Red (NIR) imaging for their potential applicability in semantic segmentation tasks. 
\par
NIR imaging has gained interest in engineering in the past few years, mostly because of the availability of sensors which could capture such images \cite{nirAnalysis}. NIR is a region (750nm-1.4$\mu$m) in electromagnetic spectrum adjacent to red. After Night Vision and Thermal cameras, NIR cameras are being seen as another extension to the range of available vision sensors. However, in order to determine the effectiveness of such cameras in various vision tasks, investigations into the matter is needed. Some of the obvious applications of NIR cameras lie in close proximity to thermal cameras \cite{nirApplication1}\cite{nirApplication2}\cite{nirApplication3} which is a low level application. An investigation into the applicability of NIR sensors for high level visual perception is not only interesting from engineering perspective rather it is an intriguing endeavor which could unearth reasons behind choices that evolutionary process has made in order to achieve the current state of visual system.

\par
The overview of the rest of the article is as follows. The Section \ref{sec:learning_frameworks} explains the methods which are used for training the models for learning object categories; Section \ref{sec:setup} explains the details of experimental setup; Section \ref{sec:related_work} enlists some of the related works; Section \ref{sec:results} presents the results obtained from experiments and Section \ref{sec:conclusions} wraps the study with some discussions and reflections. 

\section{Bio-inspired Learning Frameworks}
\label{sec:learning_frameworks}

\subsection{Convolutional Neural Network (CNN)}

\begin{figure}
      \centering
      \includegraphics[width=\columnwidth]{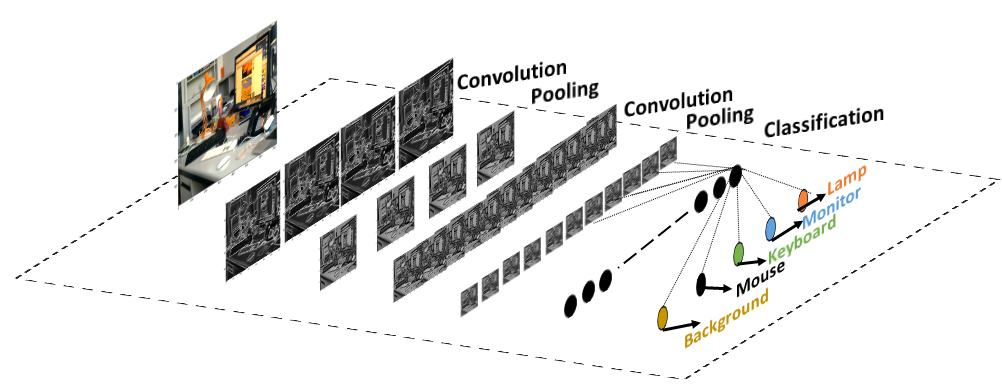}
      \caption{Example of a Convolutional Neural Network Architecture.}
      \label{fig_cnn}
\end{figure}

Convolutional Neural Network (CNN) is a feed forward multilayer perceptron with a number of convolution layers. It differs from typical neural network as it has initial set of directly connected layers which work as feature learning as well as filtering the data to lower dimension. This helps to overcome the problem of "curse of dimensionality" which is faced by a fully connected neural network and therefore enables it to scale well as the image resolution increases. Although CNN architecture can vary greatly depending on the input data as well as number of layers and their configuration, there are however, three major sections in all adaptations (See Figure \ref{fig_cnn}). First section of a CNN consists of data layers whose number depend on the dimension of the input data. The data in these layers is pre-processed with some common steps (e.g. whitening, cropping, re-sizing and mirroring). Second section consists of one or more sets of feature learning sub-section. Each feature learning sub-section consists of a convolution layer followed by a sampling layer which performs a sub-sampling of the input (e.g. max-pooling/average). Depending on the depth of the network, a number of these feature learning sections are added and a high level feature representation of the input is obtained. These high level features are then fed into a classification section of the network which consists of a set of fully connected neurons which performs as a logistic regression on the features and thus predict the probability of each class. The error is computed with respect to the known output and the error is propagated back in the network. 

\par
There has been a huge interest among the scientific community recently, in Deep Learning algorithms specially the CNNs due to their robustness and similarity to natural neural system. The application domain is quite large ranging from intelligent information processing \cite{cnn_app1} to medical applications \cite{cnn_app2} \cite{cnn_app3}. Therefore, these methods make suitable choice for answering some of the questions concerning human vision system which can lead to development of better artificial vision systems.
 
\subsection{Fully Convolutional Neural Network}

\begin{figure}
      \centering
      \includegraphics[width=\columnwidth]{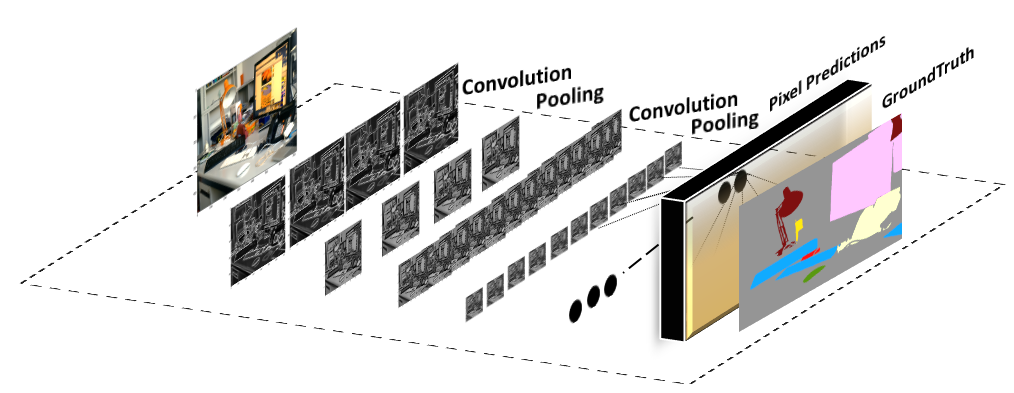}
      \caption{Example of a Fully Convolutional Neural Network Architecture.}
      \label{fig_fcn}
\end{figure}

Fully Convolutional Neural Network (FCN) is an extension to the common CNN architecture \cite{alexNet}. A CNN intends to predict probability distribution for a set of output classes which can only be useful in classifying images for their type or it can predict the presence of a target object in a candidate image. This prediction however requires a de-convolution step in order to localize the object in the input image \cite{fcn}. This localization step gives a rough location of the object in the input image and is rather useless in semantic segmentation tasks which require a precise detection of object boundaries. FCN intends to circumvent this problem by keeping a 3D format of the data in all layers of the network. It differs from the typical CNN as the latter tends to make the data linear in the fully connected part. Keeping a three dimensional representation throughout network enables to predict a 3D probability distribution consisting of a set of 2D pixel-wise predictions for each class (Figure \ref{fig_fcn}). The error is computed against the ground-truth image and pixel-wise error is computed and propagated back into the network. This kind of implementation allows FCN to accurately predict the semantic segmentation for a set of classes.

\subsection{ViperNet}

\begin{figure}
      \centering
      \includegraphics[width=\columnwidth]{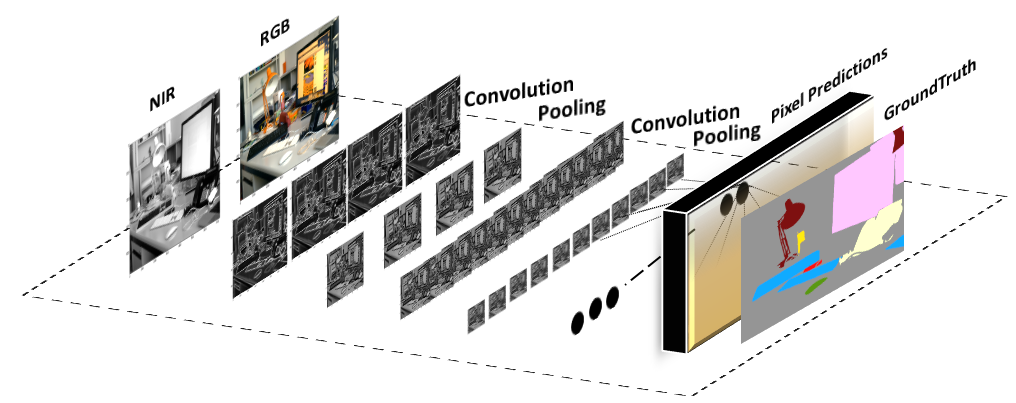}
      \caption{ViperNet Architecture.}
      \label{vcn}
\end{figure}

Since the target of the work is to experiment with NIR images, therefore, FCN has been modified to form a ViperNet which takes RGB and NIR images and generates the predictions for the semantic classification (Figure \ref{vcn}). FCN takes as input a VGG net \cite{vgg16} for initialization which consists of a 16 pixel stride. Therefore, implementation provided by FCN32s (FCN with 32 pixel stride) for 4-channel data (i.e. RGBD) is sparse which might be sufficient for depth images but is rather a coarse representation as far as NIR images are concerned. Therefore, FCN8s (FCN with 8 pixel stride) is modified to accommodate four channel data. In this way, an early integration of all four channels is performed whereby all data channels are merged before feeding to convolutional layer. 

\begin{table}
\caption{No of Instances in indoor dataset.}
\begin{center}

{\begin{tabular}{@{}lccr@{}}
\bf Class & \bf No. of Instances\\ 
\hline
BG & 400 \\
 \hline 
 Bottle & 130 \\
 \hline 
 Can & 60 \\
 \hline 
 CellPhone & 109 \\
 \hline 
 ClothingItem & 184 \\
 \hline 
 Cup & 162  \\
 \hline 
 FlowerPot & 76 \\
 \hline 
 Handbag & 124 \\
 \hline 
 Keyboard & 179 \\
 \hline 
 Mouse & 146 \\
 \hline
 OfficeLamp & 71 \\
 \hline 
 OfficePhone & 113 \\
 \hline 
 Screen & 205 \\  
\hline
\end{tabular}}
\label{indoor}
\end{center}
\end{table}

\section{Experimental Setup}
\label{sec:setup}

For all the experiments in this work a RGB-NIR dataset has been utilitzed \cite{rgbnir}. The dataset has total of 770 images which consists of 400 indoor and 370 outdoor images. The images are captured by SLR camera with NIR filter with cutoff at 750nm. The indoor images consists of 13 object classes while outdoor images consists of 10 object classes. The dataset also provides ground-truths in the form of marked objects in the images. A detail of individual object instances for each class in the dataset is given in table \ref{indoor} and table \ref{outdoor}. The images in both indoor and outdoor datasets have been split into training and test sets. More specifically, 70\% stratified class instances have been used for training set and the rest for test set. For each dataset, a FCN model has been trained for 200K iterations. For all experiments a fixed learning rate of 1e-10 has been used with momentum set to 0.9. A total of four FCN8s models has been build; two for RGB input and two for RGB-NIR input. The time taken for each iteration has been ~1 second. The memory foot print of the model has been 5.2 GB. A hexacore intel i7 PC with Nvidia Titan X GPU is used for all the experiments.

\par

The model is evaluated on test set using multiple performance metrics. A common set of metrics used in semantic segmentation tasks vary in terms of pixel accuracy and degree of overlap in the detected regions with respect to the ground-truth. If $p_i{}_j$ be the number of instances of class $i$ predicted as class $j$, where $n_c$ is the total number of classes and let $n_i = \sum_{j}p_i{}_j$ be the total number of pixels of class $i$ then performance metrics can be computed as:

\begin{itemize}
\item  mean IU: $1/n_c\sum_ip_i{}_i/(n_i+\sum_jp_j{}_i-p_i{}_i)$ \\
\item pixel accuracy: $sum_ip_i{}_i/n_i$ \\
\item mean accuracy: $1/n_c\sum_ip_i{}_i/n_i$ \\
\item frequency weighted IU:\\$(\sum_kn_k)^-1\sum_in_ip_i{}_i/(n_i+\sum_jp_j{}_i-p_i{}_i)$ \\

\end{itemize} 

\section{Related Work}
\label{sec:related_work}
Although the exact research question has not been addressed in any study as far as the author's knowledge is concerned. A closest work which evaluate the RGB-NIR images for semantic segmentation is reported in \cite{rgbnir}. The authors used a CRF based object detection framework and found that it might be partially useful for some classes. It differs from the work reported here, in the choice of method as well as the type of the inquisitive query. 
In addition to this, the closest work would be the semantic segmentation studies which utilize CNN and predict pixel-wise probabilities. Notably, SegNet reported in \cite{segNet} could be one of the recent state-of-the-art in addition to FCN \cite{fcn} which has already been explained in detail in the previous section. However, since FCN is more recent and has already claimed to be better performing therefore, SegNet has not been included in the experimentation.

\begin{table}
\caption{No of Instances in outdoor dataset. } 
\begin{center}
{\begin{tabular}{@{}lc@{\hskip 1cm}cr@{}}

\bf Class & \bf No. of Instances\\ 
\hline
 Building & 180 \\
 \hline 
 Cloud & 162 \\
 \hline 
 Grass & 160 \\
 \hline 
 Road & 109 \\
 \hline 
 Rock & 81 \\
 \hline 
 Sky & 175  \\
 \hline 
 Snow & 42 \\
 \hline 
 Soil & 79 \\
 \hline 
 Tree & 275 \\
 \hline 
 Water & 80 \\
\hline
\end{tabular} }
\label{outdoor}
\end{center}
\end{table}

\begin{table}
\caption{Comparison of classification Performance.}
\begin{center}
{\begin{tabular}{@{}C{2cm}C{2cm}C{1.5cm}C{2cm}@{}}
 \hline
 \bf  &  \bf Mean Accuracy & \bf Mean IU & \bf Frequency weighted IU\\ 
 \hline
 \bf Indoor (RGB) & 0.95 & 0.85 & 0.96\\ 
 \hline
  \bf Indoor (RGB+NIR)  & 0.94 & 0.80 & 0.94\\
 \hline 
 \bf Outdoor (RGB)  & 0.97 & 0.94 & 0.97\\
 \hline 
 \bf Outdoor (RGB+NIR)  & 0.96 & 0.93 & 0.96\\
\hline
\end{tabular} }
\label{accuracy}
\end{center}
\end{table}

\vspace{2mm}   
\section{RESULTS and Analysis}
\label{sec:results}

\begin{figure*}[ht]
\centering

\foreach \x in {1,2,3,4,5,6,7,8} 
{
\includegraphics[height=2.5cm]{images/indoor/rgb/\x}  \includegraphics[height=2.5cm]{images/indoor/gt/\x}    \includegraphics[height=2.5cm]{images/indoor/color/\x}      \includegraphics[height=2.5cm]{images/indoor/nir/\x}\\\vspace{2.5mm}
}
\includegraphics[width=\textwidth]{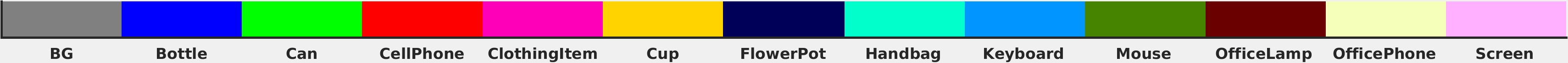}\\\vspace{2.5mm}
\caption{Results of Classification on indoor dataset. Column 1: original image, Column 2: Ground-Truth, Column 3: Results using only RGB, Column 4: Results using RGB+NIR}
\label{fig_indoor_visualization}
\end{figure*}

\begin{figure*}[ht]
\centering

\foreach \x in {1,2,3,4,5,6,7,8} 
{
\includegraphics[height=2.5cm]{images/outdoor/rgb/\x}  \includegraphics[height=2.5cm]{images/outdoor/gt/\x}    \includegraphics[height=2.5cm]{images/outdoor/color/\x}      \includegraphics[height=2.5cm]{images/outdoor/nir/\x}\\\vspace{2.5mm}
}
\includegraphics[width=\textwidth]{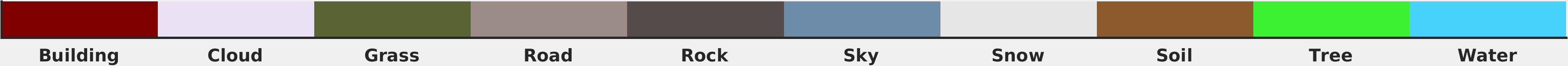}\\\vspace{2.5mm}
\caption{Results of Classification on outdoor dataset. Column 1: original image, Column 2: Ground-Truth, Column 3: Results using only RGB, Column 4: Results using RGB+NIR}
\label{fig_outdoor_visualization}
\end{figure*}

\begin{figure*}
      \subfloat[][]{\includegraphics[width=0.5\textwidth]{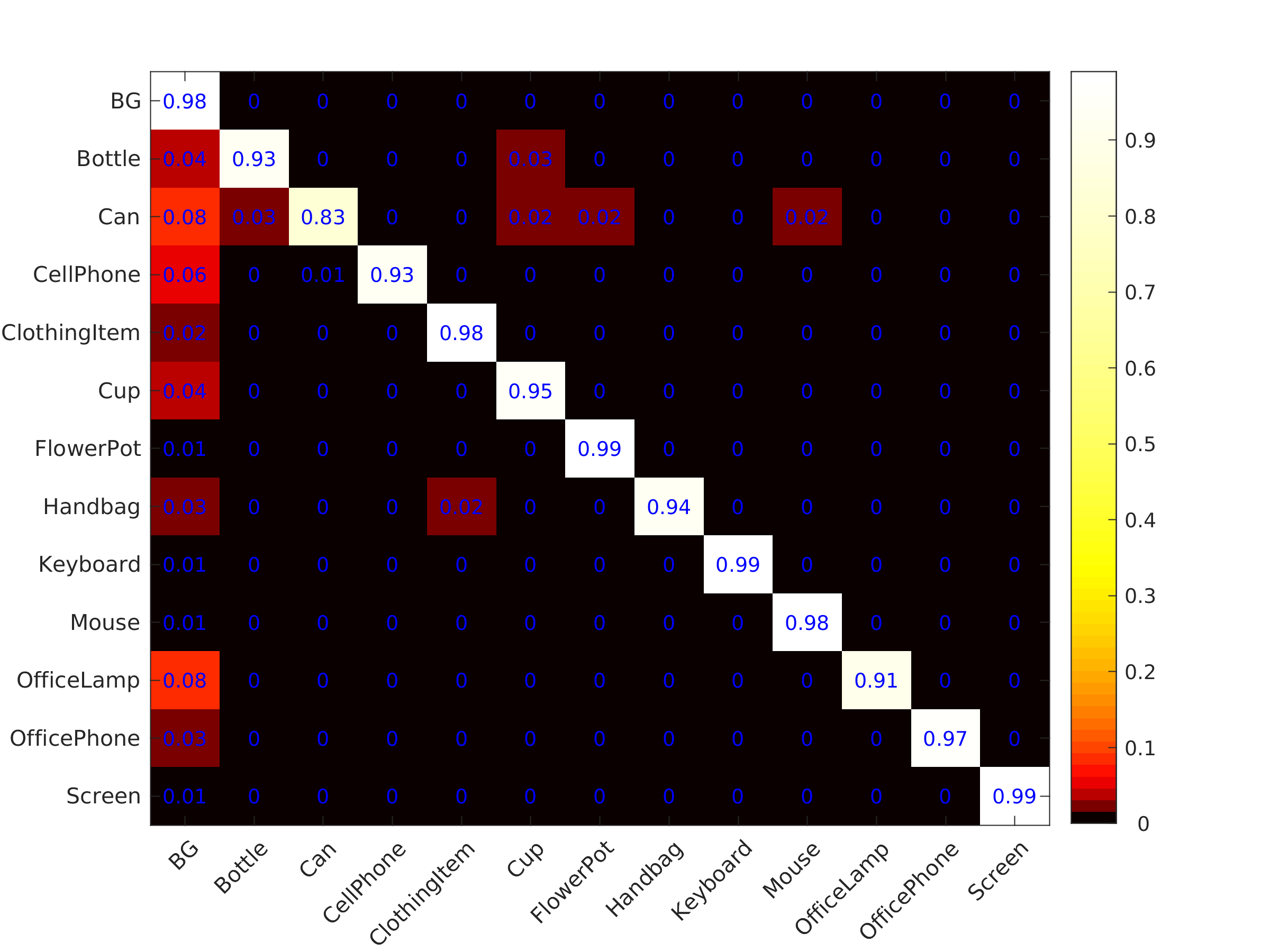}\label{fig_indoor_color_perf}}
\hfill
      \subfloat[][]{\includegraphics[width=0.5\textwidth]{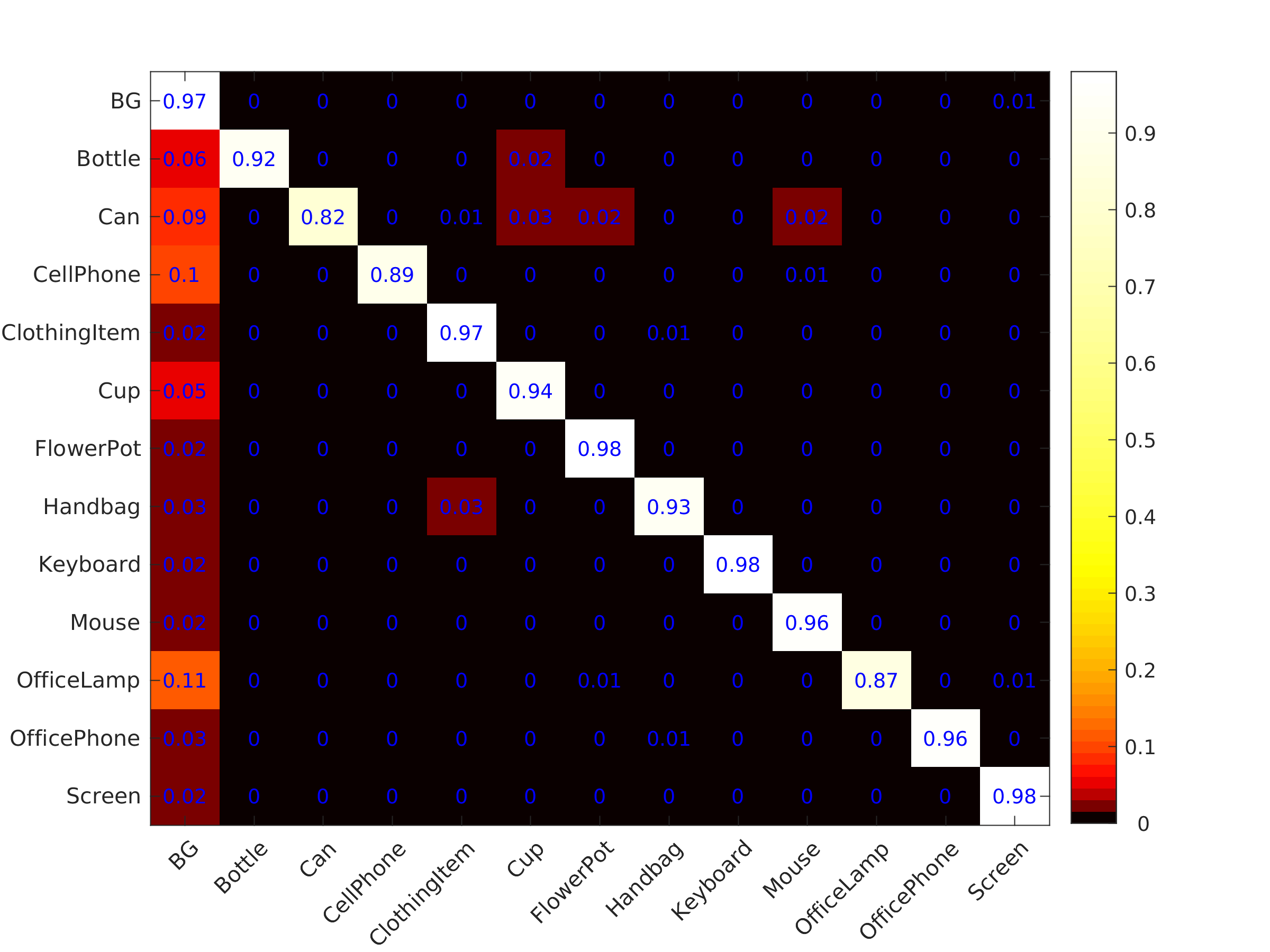}\label{fig_indoor_nir_perf}}
\hfill
      \subfloat[][]{\includegraphics[width=0.5\textwidth]{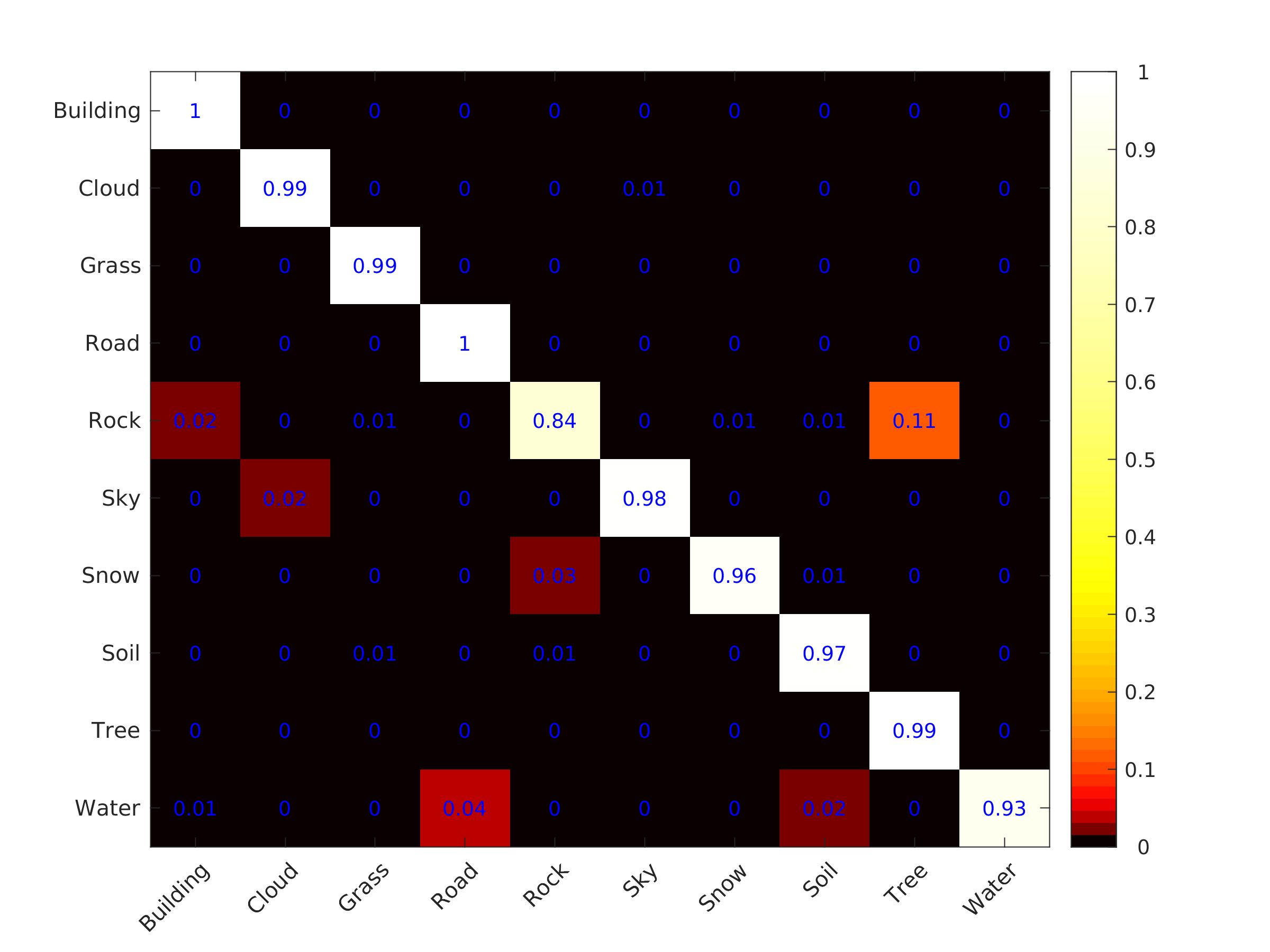}\label{fig_outdoor_color}}
\hfill
      \subfloat[][]{\includegraphics[width=0.5\textwidth]{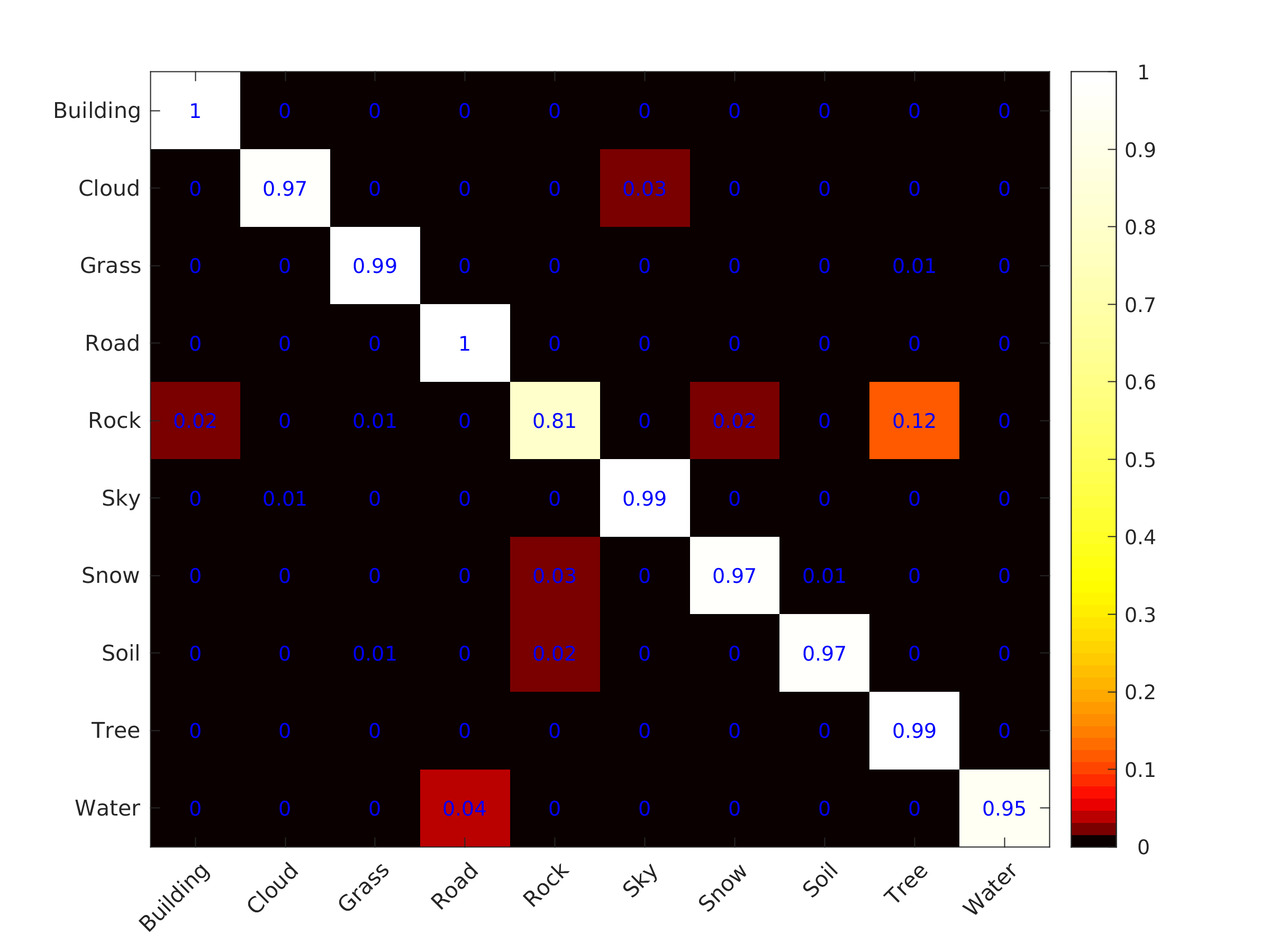}\label{fig_outdoor_color_nir}}
\hfill
      \caption{Performance Evaluation: (a) Classification of indoor images using RGB. (b) Classification of indoor images RGB+NIR. (c) Classification of outdoor images using RGB. (d) Classification of outdoor images RGB+NIR.}            
      \label{fig_performance}
\end{figure*}

We would like to know that whether evolution has favored a filtration framework in order to restrict animals within a spectral range for the benefit of high level cognitive functions such as semantic segmentation. It can be answered by comparing the two types of data inputs while keeping the methodology constant. In this respect, the experiments generated the results as given in table \ref{accuracy}. The results are pretty much similar as one would expect as the method remains the same. However, if there would be significant benefit for using RGB+NIR in this context one would expect to see some drastic differences. When we look at the qualitative results given in figure \ref{fig_indoor_visualization} and figure \ref{fig_outdoor_visualization} for indoor and outdoor datasets respectively, we notice that using extra NIR channel has rather impacted badly on multiple occasions. For example, in indoor results, background classification is often confused more with objects compared to RGB channel based results. Similarly, one could observe false positives in the outdoor results as well. This might be due to the fact that NIR channel is more susceptible to noise than mere RGB alone. 
\par
when we analyze more closely the results for the individual class performance given in figure \ref{fig_performance}, we notice that although RGB+NIR has occasional slight boost in performance in a certain class, it came on the expense of increase in false alarms. Therefore, on the whole, using only visible part of the spectrum (i.e. RGB) alone gives better performance. 
\par
If we now analyze the results of the study \cite{rgbnir} in which authors performed the classifications on the same dataset using a CRF based classifier, we see the similar pattern there as well. Although authors managed to find better performance for some classes however, overall, using visible spectrum alone, turns out to sufficient. The better performance for couple of classes can be attributed to the weak learning model. Since the method used in this work (i.e. FCN) is more recent and has overall better performance for the same dataset, it can safely be inferred that using extra NIR channel has not produced significantly better results.

\vspace{2mm}  

\vspace{2mm}  

\vspace{2mm}   
\section{Conclusions}
\label{sec:conclusions}

This work raised an interesting fundamental question which seeks to determine whether using NIR in high level vision tasks such as semantic segmentation can benefit. If it could benefit, then, it would mean that evolution didn't have any preference towards spectrum filtration which we observe when we explore vision system of the animal kingdom. However, after analyzing the results obtained from the experiments performed on a RGB-NIR dataset, it becomes clear that adding extra channel doesn't significantly improve the classification performance. Although, one thing is to be noted that the study do not claim that NIR can not be beneficial in other domains of vision, it only shows that the NIR is not much beneficial and safely be skipped when performing semantic segmentation tasks. In other words: evolution did get it right.
\\
\\

\bibliographystyle{IEEEtran}
\bibliography{ref}

\end{document}